\documentclass[11pt,a4paper]{article}
\usepackage[hyperref]{acl2020}
\usepackage{times}
\usepackage{latexsym}

\usepackage{microtype}

\usepackage{graphicx}
\usepackage{url}
\usepackage{tabularx}
\usepackage{color, colortbl}
\usepackage{cancel}
\usepackage[normalem]{ulem}
\usepackage{todonotes}

\definecolor{g4color}{HTML}{DF505A}
\definecolor{g3color}{HTML}{FCABB0}
\definecolor{g2color}{HTML}{B0D5F9}
\definecolor{g1color}{HTML}{3997EE}

\usepackage{xcolor}
\usepackage{amsmath}
\usepackage{amssymb}
\usepackage{arydshln}
\usepackage{stmaryrd}
\usepackage{enumitem}
\usepackage[T1]{fontenc}
\usepackage{soul}
\usepackage[ruled,noline,linesnumbered]{algorithm2e}
\usepackage{pifont}
\usepackage{gb4e}
\usepackage{booktabs}
\usepackage{multirow}
\usepackage[smallerops]{newtxmath}
\usepackage{nicefrac}
\usepackage{adjustbox}
\usepackage{makecell}
\usepackage[caption=false]{subfig}
\usepackage{float}
\usepackage{microtype}

\noautomath

\newcommand{\Note}[1]{}
\renewcommand{\Note}[1]{\hl{[#1]}}  
\definecolor{brilliantlavender}{rgb}{0.96, 0.73, 1.0}
\definecolor{brightgreen}{rgb}{0.4, 1.0, 0.0}
\definecolor{mikadoyellow}{rgb}{1.0, 0.77, 0.05}
\definecolor{lightblue}{rgb}{0.2, 0.512, 0.883}
\definecolor{seagreen}{rgb}{0.33, 1.0, 0.62}

\usepackage{cleveref}
\crefname{appendix}{App.}{}

\DeclareRobustCommand{\hlcyan}[1]{{\sethlcolor{cyan}\hl{#1}}}
\DeclareRobustCommand{\hlpink}[1]{{\sethlcolor{pink}\hl{#1}}}
\DeclareRobustCommand{\hlgray}[1]{{\sethlcolor{lightgray}\hl{#1}}}

\DeclareMathAlphabet{\mathcal}{OMS}{cmsy}{m}{n}
\DeclareMathAlphabet{\mathbb}{U}{msb}{m}{n}

\newcommand{\ent}{\textsc{ent}\xspace}
\newcommand{\neu}{\textsc{neu}\xspace}
\newcommand{\con}{\textsc{con}\xspace}

\newcommand{\entlabel}{\textsc{\hlcyan{ent}}\xspace}
\newcommand{\neulabel}{\textsc{\hlgray{neu}}\xspace}
\newcommand{\conlabel}{\textsc{\hlpink{con}}\xspace}

\newcommand{\p}{$p$\xspace} 
\newcommand{\h}{$h$\xspace} 

\newcommand{\unliData}{\textit{u}-SNLI\xspace}

\aclfinalcopy 
\def\aclpaperid{***} 

\title{Uncertain Natural Language Inference}

\renewcommand{\thefootnote}{\fnsymbol{footnote}}

\author{
  Tongfei Chen\textsuperscript{\rm 1}\footnotemark[1] \quad\quad Zhengping Jiang\textsuperscript{\rm 2}\footnotemark[1]\,\,\footnotemark[2] \quad\quad Adam Poliak\textsuperscript{\rm 1} \\ \bf Keisuke Sakaguchi\textsuperscript{\rm 3}\footnotemark[2] \quad\quad Benjamin Van Durme\textsuperscript{\rm 1} \\
  \textsuperscript{1}~Johns Hopkins University \\
  \textsuperscript{2}~Columbia University \\
  \textsuperscript{3}~Allen Institute for AI \\
   \texttt{\string{tongfei, azpoliak, vandurme\string}@jhu.edu,  zj2265@columbia.edu, keisukes@allenai.org}
}

\date{}

\begin{document}
\maketitle

\footnotetext[1]{~Equal contribution.}
\footnotetext[2]{~Work performed while at Johns Hopkins University.}
\renewcommand{\thefootnote}{\arabic{footnote}}

\begin{abstract}

We introduce \emph{Uncertain Natural Language Inference} (UNLI), a refinement of Natural Language Inference (NLI) that shifts away from categorical labels, targeting instead the direct prediction of subjective probability assessments. 
We demonstrate the feasibility of collecting annotations for UNLI by relabeling a portion of the SNLI dataset under a probabilistic scale, where items even with the same categorical label differ in how likely people judge them to be true given a premise.  
We describe a direct scalar regression modeling approach, and find that existing categorically labeled NLI data can be used in pre-training.  Our best models approach human performance,
demonstrating models may be capable of more subtle inferences than the categorical bin assignment employed in current NLI tasks.
\end{abstract}

\section{Introduction}

Variants of entailment tasks have been used for decades in benchmarking systems for natural language understanding.  Recognizing Textual Entailment (RTE) or Natural Language Inference (NLI) is traditionally a categorical classification problem: predict which of a set of discrete labels apply to an inference pair, consisting of a premise ($p$) and hypothesis ($h$).  The FraCaS consortium offered the task as an evaluation mechanism, along with a small challenge set~\cite{cooper1996using}, which was followed by the RTE challenges~\cite{rte-1}.
Despite differences between these and recent NLI datasets~\cite[\textit{i.a.}]{marco_marelli_2014_2787612,lai-etal-2017-natural,williams2018broad,scitail}, NLI hsa remained a categorical prediction problem.

However, \emph{entailment inference is uncertain and has a probabilistic nature}~\cite{Glickman:2005:PCA:1619499.1619502}. Maintaining NLI as a categorical classification problem is not ideal since coarse categorical labels mask the uncertain and probabilistic nature of entailment inference.
NLI pairs may share a coarse label, but the probabilities that the hypotheses are entailed by their corresponding premises may vary greatly (see \autoref{tab:float1}). Hence, not all \textit{contradictions are equally contradictory} and not all \textit{entailments are equally entailed}.

\begin{table}[t!]
  \centering
  \begin{adjustbox}{width=\linewidth}
  \begin{tabular}{lcl} 
    \toprule
    \bf Premise $\leadsto$ Hypothesis & \bf NLI & \bf UNLI \\ 
    \midrule
    A man in a white shirt taking a picture \\
    $\leadsto$ A man takes a picture & \multirow{-2}{*}{\entlabel} & \multirow{-2}{*}{100\%} \\
    A boy hits a ball, with a bat \\
    $\leadsto$ The kid is playing in a baseball game& \multirow{-2}{*}{\entlabel} & \multirow{-2}{*}{78\%} \\
    A wrestler in red cries, one in blue celebrates \\
    $\leadsto$ The wrestler in blue is undefeated & \multirow{-2}{*}{\conlabel} & \multirow{-2}{*}{50\%} \\
    Man laying on a platform outside on rocks \\
    $\leadsto$ Man takes a nap on his couch & \multirow{-2}{*}{\conlabel} & \multirow{-2}{*}{0\%} \\
  \bottomrule 
  \end{tabular}
  \end{adjustbox}
  \caption{Probability assessments on NLI pairs. The NLI and UNLI columns respectively indicate the categorical label (from SNLI) and the subjective probability for the corresponding pair.}
  \label{tab:float1}
\end{table}

We propose \emph{Uncertain Natural Language Inference} (UNLI), a refinement of NLI that captures more subtle distinctions in meaning by  shifting away from categorical labels to the direct prediction of human subjective probability assessments. We illustrate that human-elicited probability assessments contain subtle distinctions on the likelihood of a hypothesis conditioned on a premise,
and UNLI captures these distinctions far beyond categorical labels in popular NLI datasets. 

We demonstrate how to elicit UNLI annotations. Using recent large-scale language model pre-training, we provide experimental results illustrating that systems can often predict UNLI judgments, but with clear gaps in understanding. We conclude that scalar annotation protocols should be adopted in future NLI-style dataset creation, which should enable new work in modeling a richer space of interesting inferences.

\begin{table*}[t!]
\centering
\begin{adjustbox}{max width=\textwidth}
\begin{tabular}{r@{\hspace{0.2cm}}c@{\hspace{0.2cm}}lcl}
\toprule
\bf Premise & $\leadsto$ & \bf {Hypothesis } & \bf SNLI & \bf \unliData \\ \midrule
\multirow{5}{*}{A man is singing into a microphone.} &$\leadsto$& {A man performs a song.}  & \neulabel & 95\% \\ 
 &$\leadsto$& A man is performing on stage. & \neulabel & 84\% \\ 
 & $\leadsto$ & A male performer is singing a special and meaningful song. & \neulabel & 15\% \\ 
 &$\leadsto$& A man performing in a bar. & \neulabel &14\% \\ 
 &$\leadsto$& A man is singing the national anthem at a crowded stadium. & \neulabel & 0.6\% \\ \bottomrule
\end{tabular}
\end{adjustbox}
\caption{A premise in SNLI with its 5 hypotheses (labeled as neutral in SNLI)  annotated in \unliData.}
\label{tab:examples-train}
\end{table*}

\section{Eliciting UNLI annotations}

We elicit subjective probabilities from crowdsource workers (MTurk) for premise-hypothesis pairs from existing NLI data.
Annotators are asked to estimate how likely the situation described in the hypothesis sentence would be true given the premise.
Following the Efficient Annotation of Scalar Labels framework~\cite[EASL;][]{sakaguchi2018efficient}, we present annotators $5$ sentence-pairs, each with a slider bar enabling direct assessment for each pair and ask annotators to calibrate their score for a sentence-pair based on the scores they provided to the other four pairs.\footnote{~Example pairs were provided in the instructions along with suggested probability values. See \autoref{appendix:annotation} for details of the annotation interface and qualifications.}

In contrast to the uniform scale employed in the original EASL protocol, we modify the interface to allow finer-grained values near 0.0 and 1.0, following psychological findings that humans are especially sensitive to values near the ends of the probability spectrum \cite{tversky1981framing}.\footnote{~This is called the \textit{certainty effect}: more sensitivity to the difference between, e.g., 0\% and 1\% than 50\% and 51\%.} This interface decision is a \emph{key distinction} of this work contrasting  prior efforts that averaged Likert-scale (ordinal) annotations.  This allows us to capture the difference between NLI pairs that are both appropriately contradicted or entailed under NLI, but that have a perceived difference of less than 1\% probability.
 
In order to capture the sensitivity near these ends, we adopt a more fine-grained slider bar with 10,000 steps with a logistic transformation. Specifically, for raw score $x \in [0, 10000]$, we apply a scaled logistic function $f(x) = \sigma \left(\beta (x - 5000) \right)$ to re-scale the final result range to $[0, 1]$.  We ran pilots to tune $\beta$, and determine that people tend to choose much lower probability for some events even though they are just slightly less likely (e.g., just below 50\%).\footnote{~This phenomenon accords with the \textit{weighting function} in Prospect Theory \cite{kahneman1979prospect,tversky1992advances}, where people tend to downweight probabilities with around 0.4 or above.} Therefore, we use different $\beta$'s depending on the range of $[0, 0.5]$ or $(0.5, 1]$. Each sentence pair is annotated with 2- or 3-way redundancy. The individual responses are averaged to create a gold standard label for a premise-hypothesis pair.

\begin{figure}[t]
    \centering
        \includegraphics[trim={0 0.5cm 0 0cm},width=0.475\textwidth]{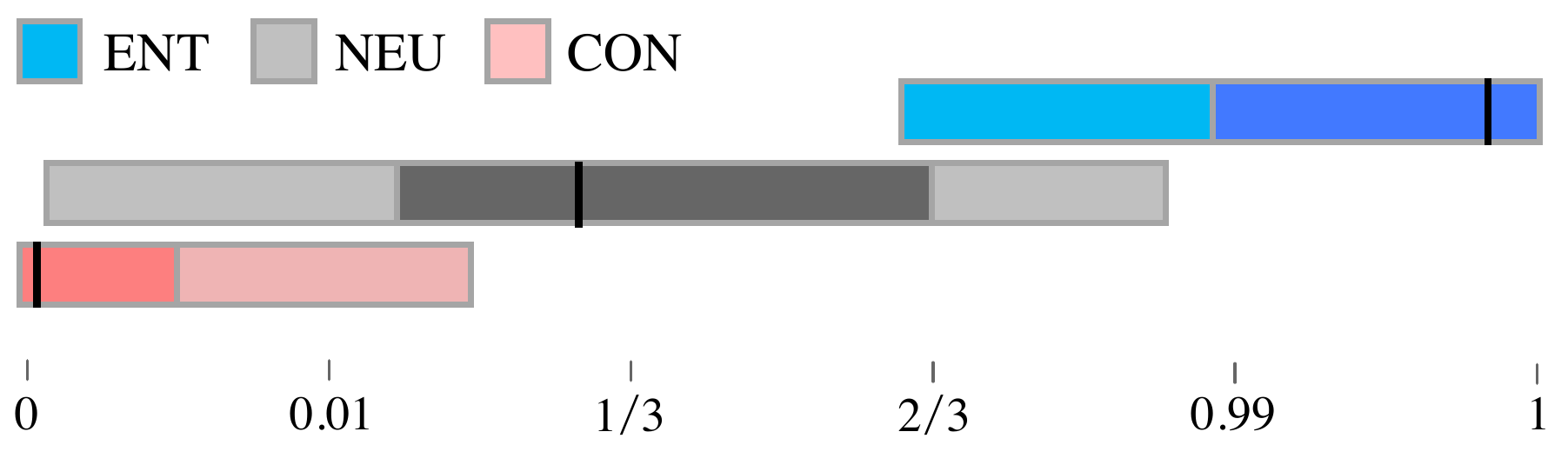} 
    \caption{Dev set statistics, illustrating median and quartile for each of the 3 categories 
    under our scalar probability scheme. Light / dark shade covers 96\% / 50\% of each category, and the bar denotes the median. Note that $x$-axis is logistic to allow fine-grained distinctions near 0.0 and 1.0.}
    \label{fig:data}
  \end{figure}

\paragraph{Data}
We annotate, i.e. elicit a probability $y \in [0, 1]$, for a subset of SNLI \cite{snli:emnlp2015} examples and refer to this data as
\unliData.\footnote{We use SNLI due to its popularity and its feature that each premise is paired with multiple hypotheses. } 
SNLI's training set contains 7,931 distinct premises paired with at least 5 distinct neutral (\neu) hypotheses. For each premise, we sample 5 neutral hypotheses, resulting in 39,655 of these \neu pairs annotated. An additional 15,862 contradicted (\con) and entailed (\ent) pairs are annotated for our training set, resulting in 55,517 training examples. For our dev and test sets, we respectively annotated 3,040 examples sampled from SNLI's dev and test splits. In total, we annotated 61,597 examples, about 12\% of all examples in SNLI.
\autoref{fig:data} plots the resultant median and quartile for each categorical SNLI label in the \unliData dev set, showing the wide range of probability judgments elicited for each label (see \autoref{tab:examples-train} for examples).\footnote{~Data is available at \url{http://nlp.jhu.edu/unli}.}

\begin{table*}[t!]
  \centering
  \begin{adjustbox}{max width=\textwidth}
  \begin{tabular}{r@{\hspace{0.2cm}}c@{\hspace{0.2cm}}lccc}
  \toprule
  \bf Premise & $\leadsto$ & \bf Hypothesis & \bf SNLI & \bf \unliData & \bf Predicted \\
  \midrule
  \makecell[r]{A man perched on a row of aquariums is using \\ a net to scoop a fish from another aquarium.} & $\leadsto$ & 	A man is standing by the aquariums. & \entlabel & 1.0 & 0.119 \\
  \midrule
  A man and woman are drinking at a bar. & $\leadsto$ & A couple is out on a date. & \neulabel & 0.755 & 0.377 \\
  Couple walking on the beach. & $\leadsto$ & The couple are holding hands. & \neulabel & 0.808 & 0.308 \\
  An elderly woman crafts a design on a loom. & $\leadsto$ & The woman is a seamstress. & \neulabel & 0.923 & 0.197 \\
  Two girls riding an amusement park ride. & $\leadsto$ & The two girls are screaming. & \neulabel & 0.909 & 0.075 \\
  \midrule
  A man and woman sit at a cluttered table.	& $\leadsto$ & The table is neat and clean. & \conlabel & 4.91$\times 10^{-4}$ & 0.262 \\
  A race car sits in the pits. & $\leadsto$ & The car is going fast. & \conlabel & 2.88$\times 10^{-7}$ & 0.724 \\
  \makecell[r]{A guy is standing in front of a toilet with a coffee \\ cup in one hand and a toilet brush in the other.} &$\leadsto$& A man is attempting to brew coffee. & \conlabel & 8.32$\times 10^{-6}$ & 0.504 \\
\bottomrule
  \end{tabular}\end{adjustbox}
  \caption{Selected \unliData dev examples where BERT predictions greatly deviate from gold assessments.}
  \label{tab:errors}
\end{table*}

\section{Prediction} 
Formally, given a premise $p \in \mathcal{P}$ and a hypothesis $h \in \mathcal{H}$, a UNLI model $F : \mathcal{P} \times \mathcal{H} \to [0, 1]$ should output an uncertainty score $\hat y \in [0, 1]$ of the premise-hypothesis pair that correlates well with a human-provided subjective probability assessment.
We train a regression UNLI model to predict the probability that a premise entails a hypothesis. We modify the sentence pair classifier\footnote{~The neural architecture for MultiNLI \cite{williams2018broad} in \citet{devlin2018bert}.} in BERT to exploit recent advancements in large-scale language model pre-training. Following \newcite{devlin2018bert}, we concatenate the premise and the hypothesis, with a special sentinel token (\textsc{cls})  inserted at the beginning and a separator (\textsc{sep}) inserted after each sentence, tokenized using WordPiece. After encoding the concatenated token sequence with BERT, we take the encoding of the first sentinel token.
\begin{equation}
    \mathbf{f}(p, h) = \mathrm{BERT}(\textsc{cls} ~;~ p ~;~ \textsc{sep} ~;~ h ~;~ \textsc{sep})[0] \ . \nonumber
\end{equation}

We pass the resulting feature vector $\mathbf{f}(p, h)$ through a sigmoid-activated linear layer to obtain a probability, instead of a softmax used in categorical NLI. 
We directly model UNLI as a \emph{regression} problem, trained using a binary cross-entropy loss\footnote{~No significant difference is observed with an $L_2$ loss.} between the human annotation $y$ and the model output $\hat{y}$. Owing to the concerns raised with \emph{annotation artifacts} in SNLI \cite{gururangan2018annotation,tsuchiya2018performance,poliak2018hypothesis}, we include a \emph{hypothesis-only baseline}.\footnote{~See \autoref{app:training-details} for additional training
details.} 

\paragraph{Metrics}
We compute Pearson correlation ($r$), the Spearman rank correlation ($\rho$), and the mean square error (MSE) between y and $\hat{y}$ as the metrics to measure the to performance of UNLI models. Pearson $r$ measures the linear correlation between the gold probability assessments and model's output; Spearman $\rho$ measures the ability of the model ranking the premise-hypothesis pairs with respect to
their subjective probability; MSE measures whether the model can recover the subjective probability value from premise-hypothesis pairs. 
A high $r$ and $\rho$, but a low MSE is desired.

\section{Results \& Analysis}

\autoref{tab:unli-training} reports results on \unliData dev and test sets. 
Just training on $55,517$ \unliData examples yields a 62.71\% Pearson $r$ on test.
The hypothesis-only baseline achieved a correlation around 40\%. This result corroborates the findings that a hidden bias exists in the SNLI dataset's hypotheses, and shows this bias may also exist in \unliData.\footnote{~This is unsurprising because \unliData examples are sampled from SNLI.} 

\begin{table}[H]
  \centering
  \begin{adjustbox}{max width=0.85\linewidth}
  \begin{tabular}{ccccc}
  \toprule
     & \multicolumn{2}{c}{\bf Hyp-only} & \multicolumn{2}{c}{\bf Full-model} \\
     \cmidrule(lr){2-3} \cmidrule(lr){4-5}
  & {\bf Dev} & {\bf Test}  & {\bf Dev} & {\bf Test}\\
  \midrule
    $\boldsymbol{r}$ & 0.3759 & 0.4120 & 0.6383 & 0.6271   \\
    $\boldsymbol{\rho}$ &0.3853 & 0.4165 &  0.6408 & 0.6346 \\ 
    {\bf MSE} & 0.1086 & 0.1055 & 0.0751 & 0.0777 \\ 
   \bottomrule
  \end{tabular}\end{adjustbox}
  \caption{Metrics for
  training on \unliData.}
  \label{tab:unli-training}
\end{table}

\paragraph{Human Performance}
We elicit additional annotations on \unliData dev set to establish a randomly sampled human performance.
We use the same annotators as before but ensure each annotator has not previously seen the pair they are annotating.
We average the scores from three-way redundant elicitation,\footnote{~This setting approximates the performance of a randomly sampled human on \unliData, and is therefore a  reasonable lower bound on the performance one could achieve with a dedicated, trained single human annotator.} yielding $r=0.6978$, $\rho=0.7273$, and $\mathrm{MSE}=0.0759$: our regression model trained on \unliData is therefore approaching human performance.  While encouraging, the model fails drastically for some examples.

\paragraph{Qualitative Error Analysis}
\autoref{tab:errors} illustrates examples with large gaps between the gold probability assessment and the BERT-based model output. The model seems to have learned lexicon-level inference (e.g., \emph{race cars} $\leadsto$ \emph{going fast}, but ignored crucial information (\emph{sits in the pits}), and fails to learn certain commonsense patterns (e.g. \emph{riding amusement park ride} $\leadsto$ \emph{screaming}; \emph{man and woman drinking at a bar} $\leadsto$ \emph{on a date}). These examples illustrate the model's insufficient commonsense reasoning and plausibility estimation. 

\paragraph{Pre-training with SNLI} 
Can we leverage the remaining roughly 500,000 SNLI training pairs that only have categorical labels? One method would be to train a categorical NLI model on SNLI and when fine-tuning on \unliData, replace the last layer of the network from a categorical prediction with a sigmoid function.\footnote{~This is similar to how \citet{babies} pre-train on SNLI, then fine-tune the model using their \emph{Add-One} pairs.} However, a typical categorical loss function would not take into account the ordering between the different categorical labels.\footnote{~That the score of \ent $>$ score of \neu $>$ score of \con.} Instead, we derive a surrogate function $s: \mathcal{T} \to [0, 1]$ that maps SNLI categorical labels $t \in \{\ent, \neu, \con\}$ to the average score of all \unliData training annotations labeled with $t$ in SNLI.\footnote{~$s: \{\ent \mapsto 0.9272; \neu \mapsto 0.4250; \con \mapsto 0.0209\}$.} 

\begin{table}[H]
  \centering
  \begin{adjustbox}{max width=0.85\linewidth}
  \begin{tabular}{ccccc}
  \toprule
     & \multicolumn{2}{c}{\bf SNLI} & \multicolumn{2}{c}{\bf SNLI + \unliData} \\
     \cmidrule(lr){2-3} \cmidrule(lr){4-5}
  & {\bf Dev} & {\bf Test}  & {\bf Dev} & {\bf Test}\\
  \midrule
    $\boldsymbol{r}$ & 0.5198 & 0.4958 & 0.6762 & 0.6589   \\
    $\boldsymbol{\rho}$ &0.5238 & 0.5231 &  0.6806 & 0.6708 \\ 
    {\bf MSE} & 0.1086 & 0.0928 & 0.0694 & 0.0733 \\ 
   \bottomrule
  \end{tabular}\end{adjustbox}
  \caption{Metrics for training only on mapped SNLI or fine-tuning on \unliData.}
  \label{tab:regression-only}
\end{table}

We use this mapping to pre-train a regression model on the SNLI training examples not included in \unliData. We also fine-tune the model on \unliData's training set. \autoref{tab:regression-only} reports the results evaluated on \unliData's dev and test sets. The model trained on the roughly $500K$ mapped SNLI examples, performs much worse than when trained on just about $55K$ \unliData examples.
When we pre-train the model on the mapped SNLI and fine-tune on \unliData, results noticeably improve. This improvement is akin to the \newcite{stilt}'s finding that many NLI datasets cover informative signal for different tasks, explaining why pre-training on NLI 
can be advantageous. Here, an impoverished version of UNLI is helpful.

\begin{figure}[H]
    \centering
    \begin{adjustbox}{width=0.48\textwidth}
      \includegraphics[trim={0.5cm 0 0 0}]
      {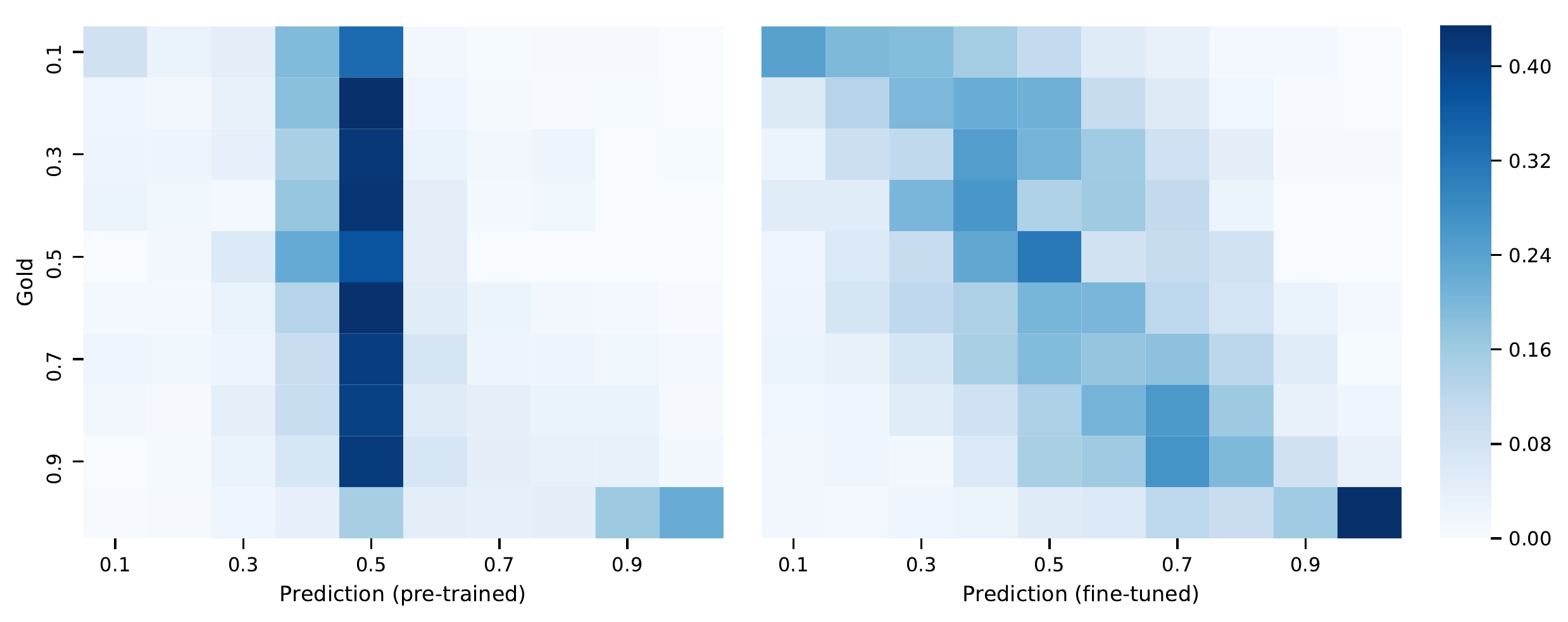}
    \end{adjustbox}
    \caption{Heatmap on \unliData dev predictions when trained only on SNLI (left) or fine-tuned on \unliData (right).
    Prediction frequencies are normalized along each gold label row.}
    \label{fig:heatmap}
\end{figure}

\paragraph{Model behavior} \autoref{fig:heatmap} depicts the model behavior when training just on SNLI or fine-tuning with \unliData. 
When using the original SNLI data, under the surrogate regression setting, the model's prediction concentrates on the 3 surrogate scalar values of the 3 SNLI classes. After fine-tuning on \unliData, the model learns smoother predictions for premise-hypothesis pairs, supported by the superior Pearson correlation score. The darker boxes in bottom-right corner of the heatmaps (\autoref{fig:heatmap}) indicate high accuracy on samples with $\approx 1.0$ gold \unliData labels and $\approx 1.0$ model predictions, signifying that our UNLI models are very good at recognizing entailments.

\section{Related Work}

The probabilistic nature and the uncertainty of NLI has been considered from a variety of perspectives. \newcite{Glickman:2005:PCA:1619499.1619502} modified the task to explicitly include the probabilistic aspect of NLI, stating that ``$p$ probabilistically entails $h$ ... if $p$ increases the likelihood of $h$ being true,'' while \newcite{lai-hockenmaier:2017:EACLlong} noted how predicting the conditional probability of one phrase given another would be helpful in predicting textual entailment. Other prior work has elicited ordinal annotations (e.g. Likert scale) reflecting likelihood judgments \cite{babies,TACL1082}, but then collapsed the annotations into coarse categorical labels for modeling. \newcite{VulicGKHK17} proposed \emph{graded lexical entailment}, which is similar to our idea but applied to lexical-level inference, asking ``to what degree $x$ is a type of $y$.'' Additionally, \citet{lalor-wu-yu:2016:EMNLP2016,lalor-EtAl:2018:EMNLP}  tried capturing the uncertainty of each inference pair by item response theory (IRT), showing fine-grained differences in discriminative power in each label.

\newcite{pavlick-tacl19} recently argued that models should ``\textit{explicitly capture the full distribution of plausible human judgments}''
as plausible human judgments cause inherent disagreements. Our concern is different as we are interested in the uncertain and probabilistic nature of NLI. We are the first to propose a method for direct elicitation of subjective probability judgments on NLI pairs and direct prediction of these scalars, as opposed to reducing to categorical classification.

Recent work have also modeled the uncertainty of other semantic phenomena  as direct scalar regression (and collected scalar versions of data for them) instead of categorical classification, e.g. factuality \cite{uw-factuality,stanovsky2017fact,neural-models-of-factuality}, and semantic proto-roles \cite{TeichertPDG17}. 

Plausiblity tasks such as COPA~\cite{roemmele2011choice} and ROCStories~\cite{mostafazadeh-EtAl:2016:N16-1} ask models to choose the most probable examples given a context, capturing \emph{relative} uncertainty between examples, but do not force a model to predict the probability of \h given \p. \citet{li-etal-2019-learning} viewed the plausibility task of COPA as a \emph{learning to rank} problem, where the model is trained to assign the highest scalar score to the most plausible alternative given context. Our work can be viewed as a variant to this, with the score being an explicit human probability judgment instead.

Linguists such as  \newcite{eijck-lappin-12}, \newcite{goodman-lassiter-2015}, \newcite{Cooper2015-COOPTT-4} and \newcite{bernardy-etal-2018-compositional} have described models for natural language semantics that introduce probabilities into the  compositional, model-theoretic tradition  begun by those such as \newcite{Davidson1967-DAVTAM-3} and \newcite{Montague1973}.  
Where they propose probabilistic models for interpreting language,  we are concerned with illustrating the feasibility of eliciting probabilistic judgments on examples through crowdsourcing, and contrasting with prior efforts restricted to limited categorical label sets.

\section{Conclusion}

We proposed \emph{Uncertain Natural Language Inference} (UNLI), a new task of directly predicting human likelihood judgments on NLI premise-hypothesis pairs. In short, we have shown that not all NLI contradictions are created equal, nor neutrals, nor entailments.
We demonstrated that (1) eliciting supporting data is feasible, and (2)  annotations in the data can be used for improving a scalar regression model beyond the information contained in existing categorical labels, using recent contextualized word embeddings, e.g. BERT.

Humans are able to make finer distinctions between
meanings than is being captured by current annotation approaches; we
advocate the community strives for systems that can do the same, and
therefore shift away from categorical NLI labels and move to something
more fine-grained such as our UNLI protocol.

\section*{Acknowledgments}
We thank anonymous reviewers from current and past versions of the article for their insightful comments and suggestions. This research benefited from support by DARPA AIDA and DARPA LORELEI. The U.S. Government is authorized to reproduce and distribute reprints for Governmental purposes. The views and conclusions contained in this publication are those of the authors and should not be interpreted as representing official policies or endorsements of DARPA or the U.S. Government.
  
\bibliography{acl2020}

\begin{thebibliography}{39}
\expandafter\ifx\csname natexlab\endcsname\relax\def\natexlab#1{#1}\fi

\bibitem[{Bernardy et~al.(2018)Bernardy, Blanck, Chatzikyriakidis, and
  Lappin}]{bernardy-etal-2018-compositional}
Jean-Philippe Bernardy, Rasmus Blanck, Stergios Chatzikyriakidis, and Shalom
  Lappin. 2018.
\newblock \href {https://www.aclweb.org/anthology/W18-4101} {A compositional
  {B}ayesian semantics for natural language}.
\newblock In \emph{Proceedings of the First International Workshop on Language
  Cognition and Computational Models}, pages 1--10. Association for
  Computational Linguistics.

\bibitem[{Bowman et~al.(2015)Bowman, Angeli, Potts, and
  Manning}]{snli:emnlp2015}
Samuel~R. Bowman, Gabor Angeli, Christopher Potts, and Christopher~D. Manning.
  2015.
\newblock \href {https://doi.org/10.18653/v1/d15-1075} {A large annotated
  corpus for learning natural language inference}.
\newblock In \emph{Proceedings of the 2015 Conference on Empirical Methods in
  Natural Language Processing}, pages 632--642.

\bibitem[{Cooper et~al.(1996)Cooper, Crouch, Van~Eijck, Fox, Van~Genabith,
  Jaspars, Kamp, Milward, Pinkal, Poesio et~al.}]{cooper1996using}
Robin Cooper, Dick Crouch, Jan Van~Eijck, Chris Fox, Johan Van~Genabith, Jan
  Jaspars, Hans Kamp, David Milward, Manfred Pinkal, Massimo Poesio, et~al.
  1996.
\newblock Using the framework.
\newblock Technical report, The FraCaS Consortium.

\bibitem[{Cooper et~al.(2015)Cooper, Dobnik, Lappin, and
  Larsson}]{Cooper2015-COOPTT-4}
Robin Cooper, Simon Dobnik, Shalom Lappin, and Stefan Larsson. 2015.
\newblock Probabilistic type theory and natural language semantics.
\newblock \emph{Linguistic Issues in Language Technology}, 10(1):1--43.

\bibitem[{Dagan et~al.(2005)Dagan, Glickman, and Magnini}]{rte-1}
Ido Dagan, Oren Glickman, and Bernardo Magnini. 2005.
\newblock \href {https://doi.org/10.1007/11736790\_9} {The {PASCAL} recognising
  textual entailment challenge}.
\newblock In \emph{Machine Learning Challenges, Evaluating Predictive
  Uncertainty, Visual Object Classification and Recognizing Textual Entailment,
  First {PASCAL} Machine Learning Challenges Workshop}, pages 177--190.

\bibitem[{Davidson(1967)}]{Davidson1967-DAVTAM-3}
Donald Davidson. 1967.
\newblock \href {https://doi.org/10.1007/BF00485035} {Truth and meaning}.
\newblock \emph{Synthese}, 17(1):304--323.

\bibitem[{Devlin et~al.(2019)Devlin, Chang, Lee, and
  Toutanova}]{devlin2018bert}
Jacob Devlin, Ming{-}Wei Chang, Kenton Lee, and Kristina Toutanova. 2019.
\newblock \href {https://doi.org/10.18653/v1/n19-1423} {{BERT:} pre-training of
  deep bidirectional transformers for language understanding}.
\newblock In \emph{Proceedings of the 2019 Conference of the North American
  Chapter of the Association for Computational Linguistics: Human Language
  Technologies, Volume 1}, pages 4171--4186.

\bibitem[{van Eijck and Lappin(2014)}]{eijck-lappin-12}
Jan van Eijck and Shalom Lappin. 2014.
\newblock Probabilistic semantics for natural language.
\newblock In Zoe Christoff, Paulo Galeazzi, Nina Gierasimczuk, Alexandru
  Marcoci, and Sonja Smets, editors, \emph{The Logic and Interactive
  Rationality Yearbook 2012}, volume~II.

\bibitem[{Glickman et~al.(2005)Glickman, Dagan, and
  Koppel}]{Glickman:2005:PCA:1619499.1619502}
Oren Glickman, Ido Dagan, and Moshe Koppel. 2005.
\newblock \href {http://dl.acm.org/citation.cfm?id=1619499.1619502} {A
  probabilistic classification approach for lexical textual entailment}.
\newblock In \emph{Proc. AAAI}, AAAI'05, pages 1050--1055. AAAI Press.

\bibitem[{Goodman and Lassiter(2015)}]{goodman-lassiter-2015}
Noah~D. Goodman and Daniel Lassiter. 2015.
\newblock Probabilistic semantics and pragmatics: Uncertainty in language and
  thought.
\newblock In Shalom Lappin and Chris Fox, editors, \emph{The Handbook of
  Contemporary Semantic Theory}, 2nd edition.

\bibitem[{Gururangan et~al.(2018)Gururangan, Swayamdipta, Levy, Schwartz,
  Bowman, and Smith}]{gururangan2018annotation}
Suchin Gururangan, Swabha Swayamdipta, Omer Levy, Roy Schwartz, Samuel~R.
  Bowman, and Noah~A. Smith. 2018.
\newblock \href {https://doi.org/10.18653/v1/n18-2017} {Annotation artifacts in
  natural language inference data}.
\newblock In \emph{Proceedings of the 2018 Conference of the North American
  Chapter of the Association for Computational Linguistics: Human Language
  Technologies, Volume 2}, pages 107--112.

\bibitem[{Kahneman and Tversky(1979)}]{kahneman1979prospect}
Daniel Kahneman and Amos Tversky. 1979.
\newblock Prospect theory: An analysis of decision under risk.
\newblock \emph{Econometrica}, 47(2):263--292.

\bibitem[{Khot et~al.(2018)Khot, Sabharwal, and Clark}]{scitail}
Tushar Khot, Ashish Sabharwal, and Peter Clark. 2018.
\newblock {SciTail}: A textual entailment dataset from science question
  answering.
\newblock In \emph{AAAI}.

\bibitem[{Kingma and Ba(2015)}]{kingma2014adam}
Diederik~P. Kingma and Jimmy Ba. 2015.
\newblock \href {http://arxiv.org/abs/1412.6980} {Adam: {A} method for
  stochastic optimization}.

\bibitem[{Lai et~al.(2017)Lai, Bisk, and Hockenmaier}]{lai-etal-2017-natural}
Alice Lai, Yonatan Bisk, and Julia Hockenmaier. 2017.
\newblock \href {https://www.aclweb.org/anthology/I17-1011/} {Natural language
  inference from multiple premises}.
\newblock In \emph{Proceedings of the Eighth International Joint Conference on
  Natural Language Processing, Volume 1}, pages 100--109.

\bibitem[{Lai and Hockenmaier(2017)}]{lai-hockenmaier:2017:EACLlong}
Alice Lai and Julia Hockenmaier. 2017.
\newblock \href {https://doi.org/10.18653/v1/e17-1068} {Learning to predict
  denotational probabilities for modeling entailment}.
\newblock In \emph{Proceedings of the 15th Conference of the European Chapter
  of the Association for Computational Linguistics, Volume 1}, pages 721--730.

\bibitem[{Lalor et~al.(2018)Lalor, Wu, Munkhdalai, and
  Yu}]{lalor-EtAl:2018:EMNLP}
John~P. Lalor, Hao Wu, Tsendsuren Munkhdalai, and Hong Yu. 2018.
\newblock \href {https://doi.org/10.18653/v1/d18-1500} {Understanding deep
  learning performance through an examination of test set difficulty: {A}
  psychometric case study}.
\newblock In \emph{Proceedings of the 2018 Conference on Empirical Methods in
  Natural Language Processing}, pages 4711--4716.

\bibitem[{Lalor et~al.(2016)Lalor, Wu, and Yu}]{lalor-wu-yu:2016:EMNLP2016}
John~P. Lalor, Hao Wu, and Hong Yu. 2016.
\newblock \href {https://doi.org/10.18653/v1/d16-1062} {Building an evaluation
  scale using item response theory}.
\newblock In \emph{Proceedings of the 2016 Conference on Empirical Methods in
  Natural Language Processing}, pages 648--657.

\bibitem[{Lee et~al.(2015)Lee, Artzi, Choi, and Zettlemoyer}]{uw-factuality}
Kenton Lee, Yoav Artzi, Yejin Choi, and Luke Zettlemoyer. 2015.
\newblock \href {https://doi.org/10.18653/v1/d15-1189} {Event detection and
  factuality assessment with non-expert supervision}.
\newblock In \emph{Proceedings of the Conference on Empirical Methods in
  Natural Language Processing, 2015}, pages 1643--1648.

\bibitem[{Li et~al.(2019)Li, Chen, and Durme}]{li-etal-2019-learning}
Zhongyang Li, Tongfei Chen, and Benjamin~Van Durme. 2019.
\newblock \href {https://doi.org/10.18653/v1/p19-1475} {Learning to rank for
  plausible plausibility}.
\newblock In \emph{Proceedings of the 57th Conference of the Association for
  Computational Linguistics}, pages 4818--4823.

\bibitem[{Marelli et~al.(2014)Marelli, Menini, Baroni, Bentivogli, Bernardi,
  and Zamparelli}]{marco_marelli_2014_2787612}
Marco Marelli, Stefano Menini, Marco Baroni, Luisa Bentivogli, Raffaella
  Bernardi, and Roberto Zamparelli. 2014.
\newblock \href {https://doi.org/10.5281/zenodo.2787612} {{The SICK (Sentences
  Involving Compositional Knowledge) dataset for relatedness and entailment}}.

\bibitem[{Montague(1973)}]{Montague1973}
Richard Montague. 1973.
\newblock \href {https://doi.org/10.1007/978-94-010-2506-5_10} {The proper
  treatment of quantification in ordinary english}.
\newblock In K.~J.~J. Hintikka, J.~M.~E. Moravcsik, and P.~Suppes, editors,
  \emph{Approaches to Natural Language: Proceedings of the 1970 Stanford
  Workshop on Grammar and Semantics}, pages 221--242. Springer Netherlands,
  Dordrecht.

\bibitem[{Mostafazadeh et~al.(2016)Mostafazadeh, Chambers, He, Parikh, Batra,
  Vanderwende, Kohli, and Allen}]{mostafazadeh-EtAl:2016:N16-1}
Nasrin Mostafazadeh, Nathanael Chambers, Xiaodong He, Devi Parikh, Dhruv Batra,
  Lucy Vanderwende, Pushmeet Kohli, and James~F. Allen. 2016.
\newblock \href {https://doi.org/10.18653/v1/n16-1098} {A corpus and cloze
  evaluation for deeper understanding of commonsense stories}.
\newblock In \emph{{NAACL} {HLT} 2016, The 2016 Conference of the North
  American Chapter of the Association for Computational Linguistics: Human
  Language Technologies}, pages 839--849.

\bibitem[{Pavlick and Callison{-}Burch(2016)}]{babies}
Ellie Pavlick and Chris Callison{-}Burch. 2016.
\newblock \href {https://doi.org/10.18653/v1/p16-1204} {Most "babies" are
  "little" and most "problems" are "huge": Compositional entailment in
  adjective-nouns}.
\newblock In \emph{Proceedings of the 54th Annual Meeting of the Association
  for Computational Linguistics, Volume 1}, pages 2164--2173.

\bibitem[{Pavlick and Kwiatkowski(2019)}]{pavlick-tacl19}
Ellie Pavlick and Tom Kwiatkowski. 2019.
\newblock \href {https://transacl.org/ojs/index.php/tacl/article/view/1780}
  {Inherent disagreements in human textual inferences}.
\newblock \emph{Trans. Assoc. Comput. Linguistics}, 7:677--694.

\bibitem[{Phang et~al.(2018)Phang, F{\'{e}}vry, and Bowman}]{stilt}
Jason Phang, Thibault F{\'{e}}vry, and Samuel~R. Bowman. 2018.
\newblock \href {http://arxiv.org/abs/1811.01088} {Sentence encoders on stilts:
  Supplementary training on intermediate labeled-data tasks}.
\newblock \emph{CoRR}, abs/1811.01088.

\bibitem[{Poliak et~al.(2018)Poliak, Naradowsky, Haldar, Rudinger, and
  Durme}]{poliak2018hypothesis}
Adam Poliak, Jason Naradowsky, Aparajita Haldar, Rachel Rudinger, and
  Benjamin~Van Durme. 2018.
\newblock \href {https://doi.org/10.18653/v1/s18-2023} {Hypothesis only
  baselines in natural language inference}.
\newblock In \emph{Proceedings of the Seventh Joint Conference on Lexical and
  Computational Semantics}, pages 180--191.

\bibitem[{Roemmele et~al.(2011)Roemmele, Bejan, and
  Gordon}]{roemmele2011choice}
Melissa Roemmele, Cosmin~Adrian Bejan, and Andrew~S. Gordon. 2011.
\newblock \href {http://www.aaai.org/ocs/index.php/SSS/SSS11/paper/view/2418}
  {Choice of plausible alternatives: An evaluation of commonsense causal
  reasoning}.
\newblock In \emph{Logical Formalizations of Commonsense Reasoning, Papers from
  the 2011 {AAAI} Spring Symposium}.

\bibitem[{Rudinger et~al.(2018)Rudinger, White, and
  Durme}]{neural-models-of-factuality}
Rachel Rudinger, Aaron~Steven White, and Benjamin~Van Durme. 2018.
\newblock \href {https://doi.org/10.18653/v1/n18-1067} {Neural models of
  factuality}.
\newblock In \emph{Proceedings of the 2018 Conference of the North American
  Chapter of the Association for Computational Linguistics: Human Language
  Technologies, Volume 1}, pages 731--744.

\bibitem[{Sakaguchi and Durme(2018)}]{sakaguchi2018efficient}
Keisuke Sakaguchi and Benjamin~Van Durme. 2018.
\newblock \href {https://doi.org/10.18653/v1/P18-1020} {Efficient online scalar
  annotation with bounded support}.
\newblock In \emph{Proceedings of the 56th Annual Meeting of the Association
  for Computational Linguistics, Volume 1}, pages 208--218.

\bibitem[{Shapiro et~al.(2014)Shapiro, Campbell, and Wright}]{bookofodds}
Amram Shapiro, Louise~Firth Campbell, and Rosalind Wright. 2014.
\newblock \href {https://books.google.com/books?id=Ayg0AgAAQBAJ} {\emph{Book of
  Odds: From Lightning Strikes to Love at First Sight, the Odds of Everyday
  Life}}.
\newblock William Morrow Paperbacks.

\bibitem[{Stanovsky et~al.(2017)Stanovsky, Eckle{-}Kohler, Puzikov, Dagan, and
  Gurevych}]{stanovsky2017fact}
Gabriel Stanovsky, Judith Eckle{-}Kohler, Yevgeniy Puzikov, Ido Dagan, and
  Iryna Gurevych. 2017.
\newblock \href {https://doi.org/10.18653/v1/P17-2056} {Integrating deep
  linguistic features in factuality prediction over unified datasets}.
\newblock In \emph{Proceedings of the 55th Annual Meeting of the Association
  for Computational Linguistics, Volume 2}, pages 352--357.

\bibitem[{Teichert et~al.(2017)Teichert, Poliak, Durme, and
  Gormley}]{TeichertPDG17}
Adam~R. Teichert, Adam Poliak, Benjamin~Van Durme, and Matthew~R. Gormley.
  2017.
\newblock \href {http://aaai.org/ocs/index.php/AAAI/AAAI17/paper/view/14997}
  {Semantic proto-role labeling}.
\newblock In \emph{Proceedings of the Thirty-First {AAAI} Conference on
  Artificial Intelligence}, pages 4459--4466.

\bibitem[{Tsuchiya(2018)}]{tsuchiya2018performance}
Masatoshi Tsuchiya. 2018.
\newblock \href
  {http://www.lrec-conf.org/proceedings/lrec2018/summaries/786.html}
  {Performance impact caused by hidden bias of training data for recognizing
  textual entailment}.
\newblock In \emph{Proceedings of the Eleventh International Conference on
  Language Resources and Evaluation}.

\bibitem[{Tversky and Kahneman(1981)}]{tversky1981framing}
Amos Tversky and Daniel Kahneman. 1981.
\newblock The framing of decisions and the psychology of choice.
\newblock \emph{Science}, 211(4481):453--458.

\bibitem[{Tversky and Kahneman(1992)}]{tversky1992advances}
Amos Tversky and Daniel Kahneman. 1992.
\newblock Advances in prospect theory: Cumulative representation of
  uncertainty.
\newblock \emph{Journal of Risk and uncertainty}, 5(4):297--323.

\bibitem[{Vuli\'{c} et~al.(2017)Vuli\'{c}, Gerz, Kiela, Hill, and
  Korhonen}]{VulicGKHK17}
Ivan Vuli\'{c}, Daniela Gerz, Douwe Kiela, Felix Hill, and Anna Korhonen. 2017.
\newblock \href {https://doi.org/10.1162/COLI\_a\_00301} {Hyperlex: {A}
  large-scale evaluation of graded lexical entailment}.
\newblock \emph{Computational Linguistics}, 43(4).

\bibitem[{Williams et~al.(2018)Williams, Nangia, and
  Bowman}]{williams2018broad}
Adina Williams, Nikita Nangia, and Samuel~R. Bowman. 2018.
\newblock \href {https://doi.org/10.18653/v1/n18-1101} {A broad-coverage
  challenge corpus for sentence understanding through inference}.
\newblock In \emph{Proceedings of the 2018 Conference of the North American
  Chapter of the Association for Computational Linguistics: Human Language
  Technologies, Volume 1}, pages 1112--1122.

\bibitem[{Zhang et~al.(2017)Zhang, Rudinger, Duh, and Durme}]{TACL1082}
Sheng Zhang, Rachel Rudinger, Kevin Duh, and Benjamin~Van Durme. 2017.
\newblock \href {https://transacl.org/ojs/index.php/tacl/article/view/1082}
  {Ordinal common-sense inference}.
\newblock \emph{Trans. Assoc. Comput. Linguistics}, 5:379--395.

\end{thebibliography}
\bibliographystyle{acl_natbib}

\clearpage
\newpage

\appendix
\section{Annotation}
\label{appendix:annotation}
Here we include information about the qualifications used to 
vet annotators. We also include screenshots of the interface used to collect annotations.

\subsection{Qualification Test}
\label{appendix:anno-qual}

  Annotators were given a qualification test to ensure non-expert workers were able to give reasonable subjective probability estimates. We first extracted seven statements from \textit{Book of Odds} \cite{bookofodds}, and manually split the statement into a bleached premise and hypothesis. We then wrote three easy premise-hypothesis pairs with definite probabilities like ($p$ = ``\textit{A girl tossed a coin.}'', $h$ = ``\textit{The coin comes up a head.}'',  probability: 0.5). We qualify users that meet both criteria: (1) For the three easy pairs, their annotations had to fall within a small error range around the correct label $y$, computed as $\delta = \frac{1}{4} \min \{y, 1-y \}$. (2) Their overall annotations have a Pearson $r > 0.7$ and Spearman $\rho > 0.4$. This qualification test led to a pool of 40 trusted annotators, which were employed for the entirety of our dataset creation.
  
\subsection{Annotation Interface}
\begin{figure}[t]
    \centering
        \includegraphics[trim={0 0 3cm 0},clip,width=\linewidth]{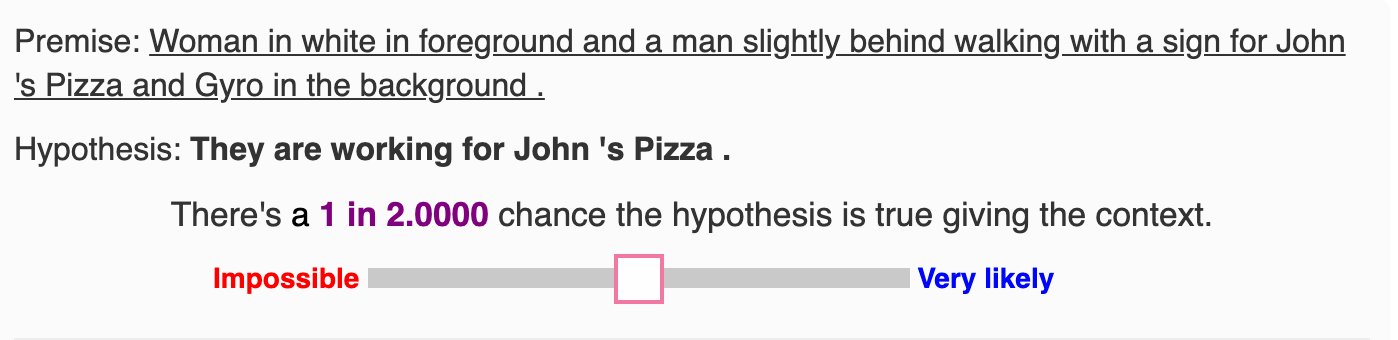}
    \caption{An example of our annotation interface.}
    \label{fig:interface}
  \end{figure}

\begin{figure}
    \centering
        \includegraphics[trim={0 0 3cm 0},clip,width=\linewidth]{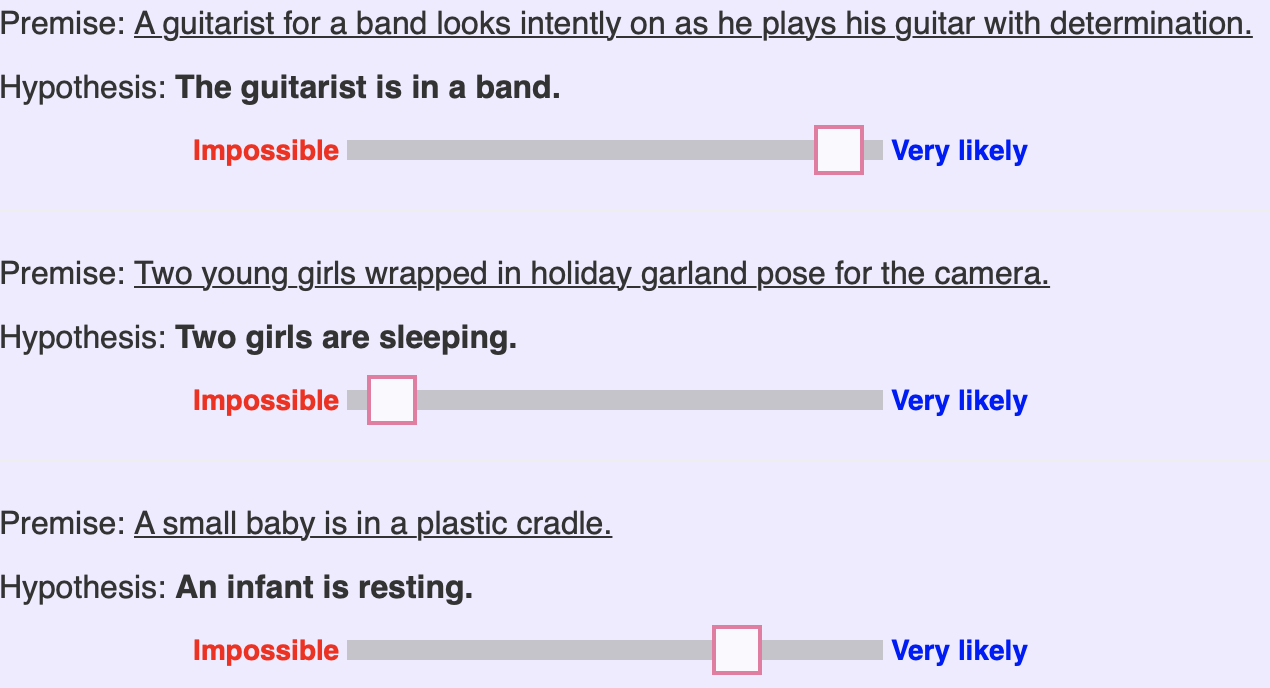}
    \caption{Three examples from the instructions.}
    \label{fig:examples}
  \end{figure}

We include screenshots of the instructions and examples shown to crowdsource workers (~\autoref{fig:examples}) as the interface we provided (\autoref{fig:interface})
  
  \section{Redundant Annotations}
  By default, we use two crowdsource workers to annotate each UNLI sentence-pair. If the two annotations on the raw slider bar $\{0, \cdots, 10000\}$ differ by more than 2000, we then elicit a third annotator.

  \section{Dataset Statistics}
  \label{app:data}
  \autoref{tab:usnli} summarizes the statistics of \unliData.
  
\section{Additional Training Details}
\label{app:training-details}
We use the \textsc{bert-base-uncased} model, with the Adam optimizer \cite{kingma2014adam}, an initial learning rate of $10^{-5}$, and maximum gradient norm 1.0. 
Our model is trained for 3 epochs, where the epoch resulting in the highest Pearson $r$ on the 
dev set is selected. 

    \begin{figure}[t]
    \centering
    \includegraphics[trim={0 0.5cm 0 0},width=\linewidth]{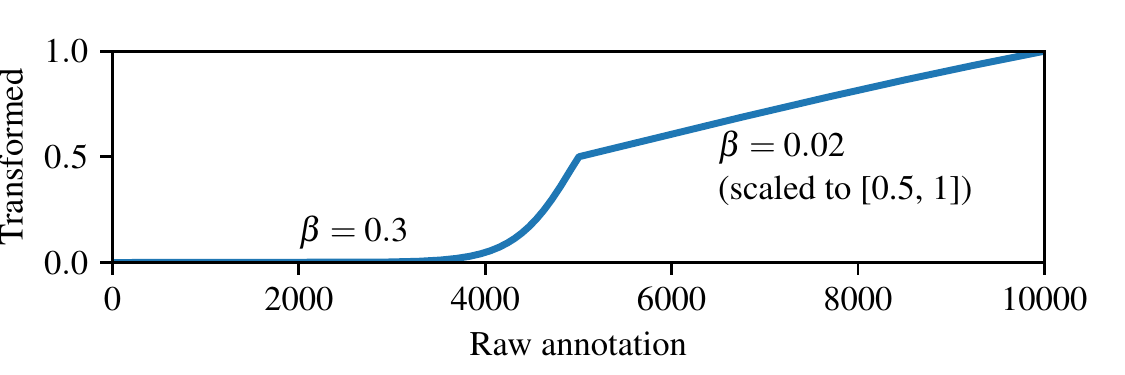}
    \caption{Our logistic transformation function. }
    \label{fig:logistic}
  \end{figure}

  \begin{table}[t]
    \centering
    \begin{adjustbox}{max width=0.45\textwidth}
    \begin{tabular}{ccrr}
    \toprule
      \bf Partition             & \bf Breakdown   & \bf SNLI & \bf U-SNLI \\
    \midrule
      \multirow{5}{*}{\bf train} & Distinct premises & 151k     & 7,931   \\
                                 & \ent hypotheses   & 183k     & 7,931   \\
                                 & \neu hypotheses   & 183k     & 39,655  \\
                                 & \con hypotheses   & 183k     & 7,931   \\
                                 & Total P-H pairs   & 550k     & 55,517  \\
    \midrule
      \multirow{5}{*}{\bf dev}   & Distinct premises & 3,319    & 2,647   \\
                                 & \ent hypotheses   & 3,329    & 162     \\
                                 & \neu hypotheses   & 3,235    & 2,764   \\
                                 & \con hypotheses   & 3,278    & 114     \\
                                 & Total P-H pairs   & 10k      & 3,040   \\
    \midrule
      \multirow{5}{*}{\bf test}  & Distinct premises & 3,323     & 2,635  \\
                                 & \ent hypotheses   & 3,368     & 156    \\
                                 & \neu hypotheses   & 3,219     & 2,770  \\
                                 & \con hypotheses   & 3,237     & 114    \\
                                 & Total P-H pairs   & 10k       & 3,040  \\
    \bottomrule
    \end{tabular}
    \end{adjustbox}
    \caption{Statistics of SNLI data re-annotated under UNLI.} 
    \label{tab:usnli}
\end{table}

\end{document}